\documentclass[letterpaper, 10 pt, conference]{ieeeconf}

\IEEEoverridecommandlockouts                              

\usepackage{amsmath}
\usepackage{amssymb}
\usepackage[colorinlistoftodos]{todonotes}
\usepackage{hyperref}
\usepackage{multirow}
\usepackage{booktabs}
\usepackage[colorinlistoftodos]{todonotes}
\usepackage{siunitx}
\usepackage{url}
\usepackage{import,bm}
\usepackage[super]{nth}
\usepackage{romannum}
\usepackage{accents}
\usepackage{verbatim}
\usepackage{tikz}
\usetikzlibrary{automata, positioning, arrows, shapes.geometric, arrows.meta, positioning, calc, intersections}
\usepackage{tkz-euclide}
\usepackage{subcaption}
\usepackage{graphicx}
\usepackage{cite}
\usepackage{makecell}
\usepackage{caption}
\usepackage{ragged2e}

\newcommand{\mt}[1]{\bm{#1}}

\title{Virtual Physical Coupling of Two Lower-Limb Exoskeletons
}

\author{Emek Barış Küçüktabak$^{1,2}$, Yue Wen$^{1}$, Matthew Short$^{1,3}$, Efe Demirbaş$^{2}$, Kevin Lynch$^{2}$, and Jose Pons$^{1,2,3}$
\thanks{$^1$ Legs and Walking Lab of Shirley Ryan AbilityLab, Chicago, IL, USA}
\thanks{$^2$ Center for Robotics and Biosystems of Northwestern University, Evanston, IL, USA}
\thanks{$^3$ Department of Biomedical Engineering, Northwestern University, Evanston, IL, USA}
}

\usepackage{atbegshi,picture}
\usepackage{lipsum}

\AtBeginShipoutNext{\AtBeginShipoutUpperLeft{%
  \put(\dimexpr\paperwidth-1cm\relax,-1.5cm){\makebox[0pt][r]{\framebox{Accepted for publication at the IEEE International Conference on Rehabilitation Robotics (ICORR), 2023}}}%
}}

\begin{document}
\maketitle
\thispagestyle{empty}
\pagestyle{empty}

\begin{abstract}
Physical interaction between individuals plays an important role in human motor learning and performance during shared tasks. Using robotic devices, researchers have studied the effects of dyadic haptic interaction mostly focusing on the upper-limb. Developing infrastructure that enables physical interactions between multiple individuals' lower limbs can extend the previous work and facilitate investigation of new dyadic lower-limb rehabilitation schemes.

We designed a system to render haptic interactions between two users while they walk in multi-joint lower-limb exoskeletons. 
Specifically, we developed an infrastructure where desired interaction torques are commanded to the individual lower-limb exoskeletons based on the users' kinematics and the properties of the virtual coupling.
In this pilot study, we demonstrated the capacity of the platform to render different haptic properties (e.g., soft and hard), different haptic connection types (e.g., bidirectional and unidirectional), and connections expressed in joint space and in task space. 
With haptic connection, dyads generated synchronized movement, and the difference between joint angles decreased as the virtual stiffness increased.
This is the first study where multi-joint dyadic haptic interactions are created between lower-limb exoskeletons.
This platform will be used to investigate effects of haptic interaction on motor learning and task performance during walking, a complex and meaningful task for gait rehabilitation.

\end{abstract}
\section{Introduction}
\label{sec:Introduction}

During shared motor tasks, humans physically interact with one another by exchanging forces to coordinate their movements. A common example of this occurs during physical therapy, where clinicians provide physical support to patients with sensorimotor impairments in order to assist them in regaining lost motor functions (e.g., guiding paretic limb foot-placement during post-stroke gait therapy). With recent advancements in human-robot interfaces, various aspects of complex physical behaviors have been explored by using robotic devices to render virtual physical environments (e.g., spring-damper) between two or more individuals~\cite{Kucuktabak2021}. With this approach, individuals can interface with separate robotic end effectors and move independently with one or more degrees of freedom (DoF) while experiencing forces based on their position and/or velocity relative to a partner's. 

\textcolor{black}{Considering the positive outcomes observed in assistive and resistive training methods~\cite{Basalp2021}, collaborative and competitive physical dyadic interaction has great potential for improved task performance and individual motor learning~\cite{Kucuktabak2021}.} Upper-limb studies in this field have shown that individuals are able to track sinusoidal trajectories more accurately while haptically coupled with a partner, and that these improvements are dependent on the skill level~\cite{Ganesh2014,Takagi2017,Beckers2020} and the properties of the virtual connection~\cite{Takagi2018}. Moreover, some findings indicate that training while haptically coupled to a partner through a compliant connection improves the rate of individual learning after the coupling is removed, in the context of 1-DoF position tracking~\cite{Ivanova2022-ui} as well as multi-DoF adaptations to visuomotor rotations~\cite{Ganesh2014} and force fields~\cite{Batson2020-wa}.
Though further investigation is warranted to generalize findings across motor tasks, studying human-human physical interaction has great promise in the development of more naturalistic controllers and interventions for individuals with neurological impairments. 

Most work involving haptic coupling between individuals has focused on upper-limb tasks, and there is limited evidence of similar behaviors in the lower limbs, despite the clear need for improved control methods in lower-limb robotic therapy~\cite{Diaz2011-bf}. Our group was the first to explore haptic interactions between pairs of individuals (dyads) performing 1-DoF ankle movements, developing infrastructure for different interactive scenarios (e.g., collaboration, competition) and coupling properties (e.g., soft, hard), as described in Kim et al.~\cite{Kim2020}. In an initial study involving a sinusoidal tracking task, we found that healthy individuals improved their tracking accuracy during compliant haptic coupling with a partner, and that these improvements were linearly related to the abilities of each partner (i.e., tracking with a better partner resulted in greater improvements)~\cite{Kim2022}. Additionally, we observed no difference in short-term, individual learning rates when comparing training alone or with a partner, consistent with the aforementioned upper-limb studies~\cite{Ivanova2022-ui,B2018,Beckers2020}, indicating some similarity between upper- and lower-limb responses to human-human interaction.

One limitation of the lower-limb infrastructure mentioned above is that the system can only support evaluation of seated, single joint movements which may lack translation to ambulatory activities critical for a healthy lifestyle~\cite{Middleton2015-ux}. In physical rehabilitation for lower-limb motor impairments associated with stroke, for instance, there is a focus on multi-joint coordination, usually through repetitive gait training supported by a therapist and/or robotic device in order to gradually achieve kinematics and muscle activation patterns which resemble unimpaired gait~\cite{Bruni2018-bj}. With this in mind, we have developed a control scheme for floating-base lower-limb exoskeletons with feet, capable of interaction torque control during walking~\cite{Kucuktabak2023haptic}. This approach also allows us to render virtual physical coupling between two or more individuals wearing these devices. In this paper, we demonstrate the feasibility of rendering virtual spring and damper elements between two individuals during treadmill walking. To the best of our knowledge, this is the first study where virtual physical coupling is created between two multi-DoF lower-limb exoskeletons during walking.

\section{Methods}
\label{sec:Methods}
\subsection{System Description}
\begin{figure}[t!]
\centering
\includegraphics[width = 0.6\columnwidth]{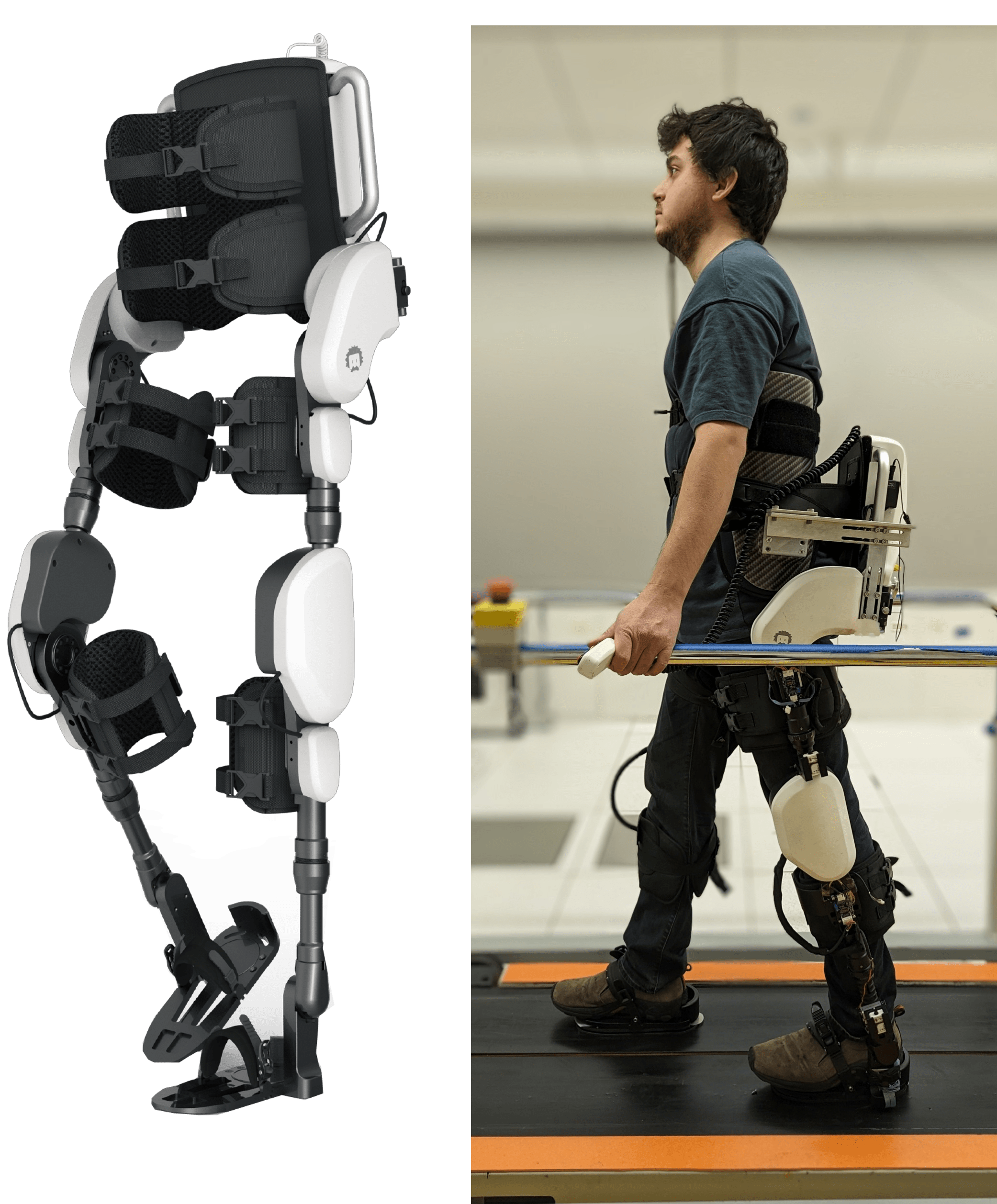}
\caption{ExoMotus-X2 lower-limb exoskeleton (left) and its modified version with a user (right).}
\label{fig:x2}
\end{figure}

Two lower-limb exoskeletons (ExoMotus-X2, Fourier Intelligence, Singapore), shown in Figure~\ref{fig:x2}, were adapted and used as a platform to test the proposed dyadic physical interaction scheme. The X2 exoskeleton is designed for overground walking, has four total active DoF at the hip and knee joints, allows passive motion at the ankle joints, and has feet that help transfer its weight to the ground.

The exoskeletons were modified to include an IMU on the backpack to measure the angular orientation and four strain gauges at each of the hip and knee joints to measure torques. Furthermore, one exoskeleton was equipped with a custom sensorized sole with 16 force sensitive resistors at the bottom of the foot. The other exoskeleton was used on a sensorized treadmill with eight 3-DoF force plates (9047B, Kistler). Measurements from these sensors were used to estimate the continuous gait states.

Communication with motors and onboard sensors was established over CAN bus using the CANOpen communication protocol. To minimize communication delay and allow real-time visualization, an external PC was connected to both exoskeletons by tethers. The controllers were implemented on a ROS and C++ based open-source software stack called CANOpen Robot Controller (CORC)~\cite{FongCanOpenDevelopment} and were run at 333 Hz on the external PC.

A Unity application was created to provide visual feedback of a reference trajectory for the ankle position and to visualize the state of the user. The communication between the CORC and the Unity application was obtained through ROS serialized messages. This visual feedback through Unity application was available only to user A.

\begin{figure}[t!]
\centering
\vspace*{0.2cm}\includegraphics[width = 1.0\columnwidth]{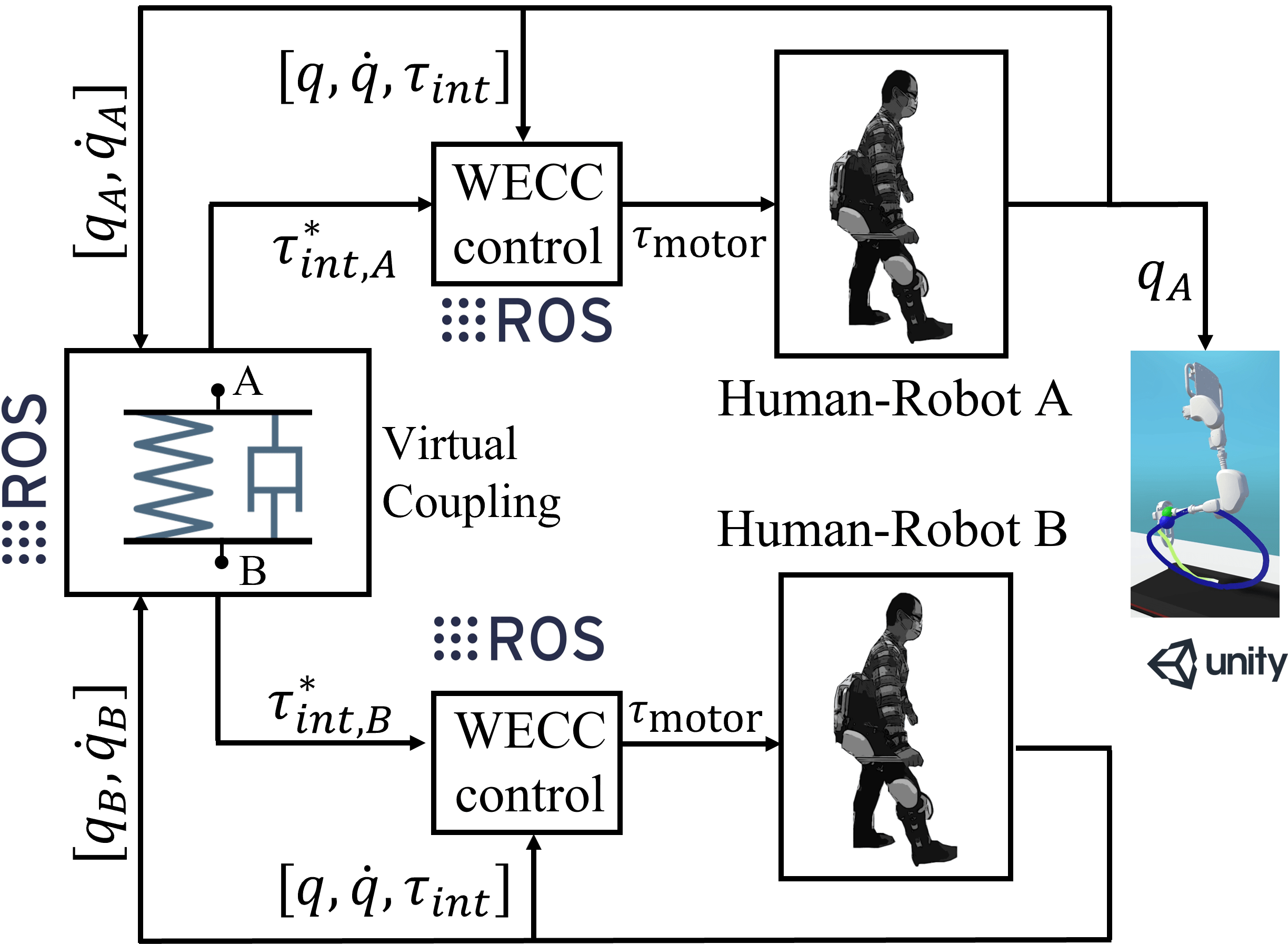}
\caption{Overall schematic of the dyadic interaction.}
\label{fig:dyad_scheme}
\end{figure}

Two separate instances of the whole-exoskeleton closed-loop compensation (WECC) controllers~\cite{Kucuktabak2023haptic} receive the desired interaction torque values, which are commanded by the \emph{virtual coupling} node. This node subscribes to the instantaneous states of both exoskeletons and sends the desired interaction torques based on the properties of the virtual physical coupling. The overall communication scheme is shown in Figure~\ref{fig:dyad_scheme}.

 WECC control compensates for the whole-body dynamics of the exoskeleton and can render zero or nonzero interaction torques. In this control loop, interaction torques between the human and exoskeleton are calculated using the joint torque measurements during the whole gait cycle. Then, with a virtual model controller, interaction torque errors are converted to desired acceleration commands. Desired acceleration commands are fed into a constrained optimization scheme to track the reference values under physical and safety limitations. Interested readers may refer to \cite{Kucuktabak2023haptic} for the details of this interaction torque controller and the whole-body model of the used exoskeleton.

\subsection{Physical Dyadic Interaction}

The main characteristics of the virtual coupling can be divided into three groups: (1) properties of the virtual coupling (e.g., stiffness, damping), (2) interaction space (i.e., joint space, task space), and (3) directionality (i.e., unidirectional, bidirectional). In this validation study, we focus on rendering virtual spring and damper elements with different characteristics.

\subsubsection{Interaction in joint space}

Interaction in the joint space is created by commanding interaction torques based on differences in the instantaneous positions and velocities of the users' joints. Desired interaction torques for the bidirectional case are 
\begin{equation} \label{eq:joint_space}
\begin{split}
    \mt{\tau}_\text{int}^\text{A} &= \mt{K}(\mt{q_\text{A}} - \mt{q_\text{B}} - \mt{q^0}) + \mt{C}(\dot{\mt{q}}_\text{A} - \dot{\mt{q}}_\text{B}) \\
    \mt{\tau}_\text{int}^\text{B} &= - \mt{\tau}_\text{int}^\text{A},
\end{split}
\end{equation}
where $\mt{K}\in \mathbb{R}^{4\times 4}$ and $\mt{C}\in \mathbb{R}^{4\times 4}$ are the diagonal torsional stiffness and damping matrices, respectively. The variables $\mt{q_\text{A, B}}\in \mathbb{R}^{4}$ represent the hip and knee joint angles from both legs of user A and B. The neutral angles of the spring are shown by the vector $\mt{q^0} \in \mathbb{R}^{4}$.

\subsubsection{Interaction in task space}
In addition to rendering torsional elements between the joints, it is also possible to render linear physical elements between any two points of the users on their lower limbs. To demonstrate this, we rendered virtual linear spring and dampers between the swing ankle positions of the users in the sagittal plane.

The force due to the virtual elements for the bidirectional case is 
\begin{equation} \label{eq:task_space_force}
\begin{split}
    \mt{F}_\text{int}^\text{A} &= \mt{K}(\mt{r_\text{A}} - \mt{r_\text{B}} - \mt{r^0}) + \mt{C}(\dot{\mt{r}}_\text{A} - \dot{\mt{r}}_\text{B}) \\
     \mt{F}_\text{int}^\text{B} &= -  \mt{F}_\text{int}^\text{A},
\end{split}
\end{equation}
where $\mt{r_\text{A,B}} \in \mathbb{R}^{2}$ is the swing ankle's horizontal and vertical position expressed in a coordinate frame attached to the stance ankle which is assumed to be fixed on the ground. $\mt{r_\text{A,B}}$ are functions of hip and knee joint positions and angle of backpack with respect to the gravity vector. Horizontal and vertical stiffness and damping are expressed with diagonal $\mt{K}\in \mathbb{R}^{2\times 2}$ and $\mt{C}\in \mathbb{R}^{2\times 2}$ matrices, respectively. The neutral horizontal and vertical length of the spring are represented by  $\mt{r^0} \in \mathbb{R}^{2}$.

The resultant torques determined by the virtual linear forces on the ankle are calculated and sent to the interaction torque controllers as shown in Equation~\ref{eq:task_space_torque}.
\begin{equation}
    \mt{\tau}_\text{int}^\text{i} = \mt{J_{\text{i}}^T} \mt{F}_\text{int}^\text{i}, \;\: i\in \{\text{A}, \text{B}\},
    \label{eq:task_space_torque}
\end{equation}
where $\mt{J_{\text{i}}^T} \in  \mathbb{R}^{5\times 2}$ is the Jacobian of the swing ankle expressed in the coordinate frame of the stance ankle. Note that this results in five-dimensional desired interaction torques, which include the backpack as well as the hip and knee joints. Moreover, even though the virtual connection is created between the swing ankles, the effect of the  connection is felt on all joints including the stance leg.

\subsubsection{Unidirectional interaction}

In clinical scenarios where a therapist needs to assess the capabilities of a patient or demonstrate a desired motion, it might be beneficial to implement this virtual physical coupling as unidirectional (i.e., only one user feels forces due to the virtual coupling). In such cases, $\mt{\tau}_\text{int}^\text{A}$ and $\mt{F}_\text{int}^\text{A}$ in Equations~\eqref{eq:joint_space} and~\eqref{eq:task_space_force}, respectively, can be set to zero to provide haptic transparency to one of the users.

\subsection{Validation Setup}

A preliminary validation experiment was conducted on an able-bodied dyad to evaluate the performance and technical feasibility of the proposed framework. \textcolor{black}{The demonstration of the infrastructure is also shown in this video\footnote{\url{https://tinyurl.com/ExoDyad}}.}

\begin{figure}[t!]
\centering
\includegraphics[width = 0.8\columnwidth]{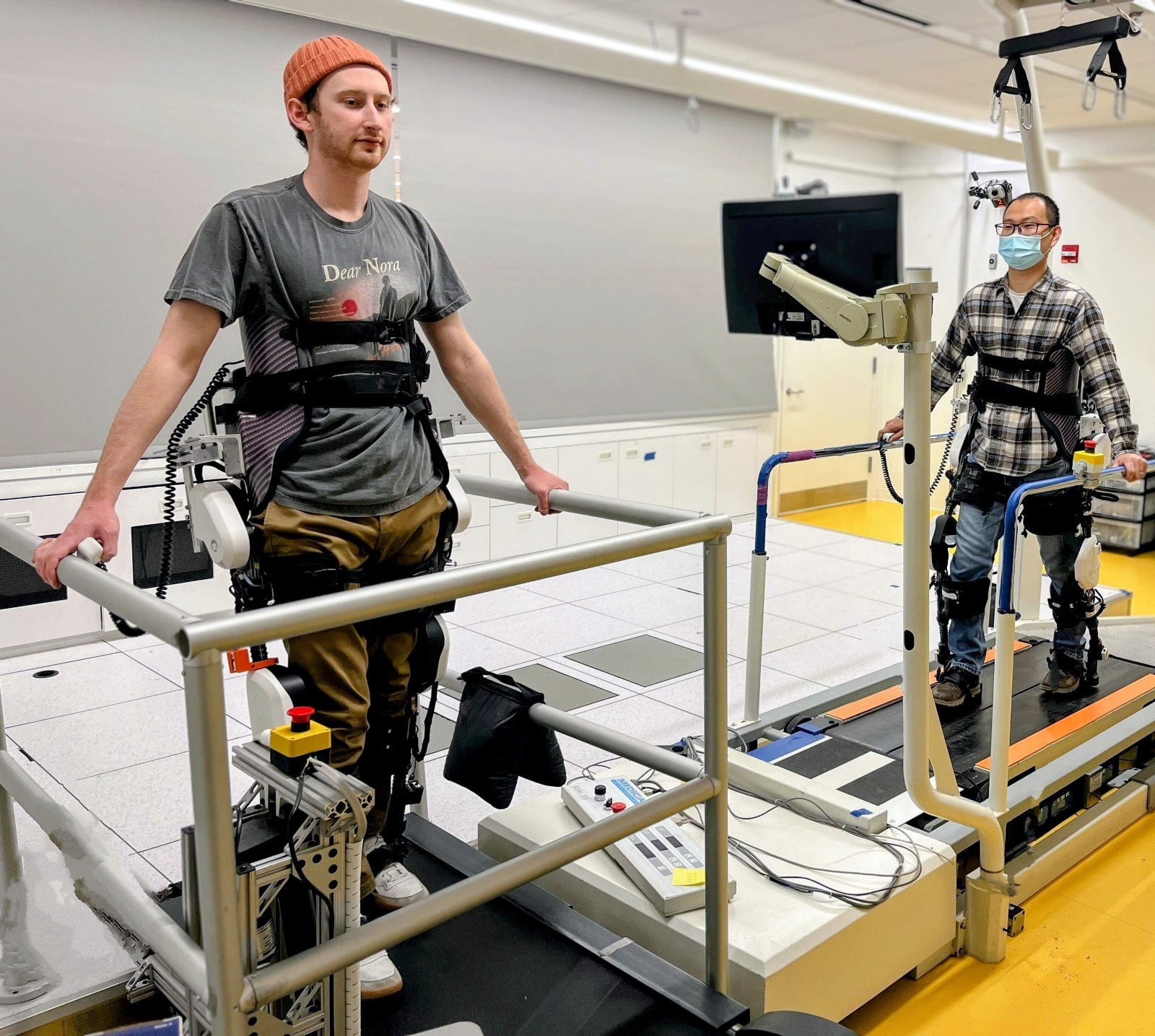}
\caption{Validation setup with a dyad.}
\label{fig:experimental_setup}
\end{figure}

Two users walked on treadmills simultaneously under different virtual physical connections. In addition, user A received visual feedback displayed on a monitor in some trials. The experimental setup is shown in Figure~\ref{fig:experimental_setup}.

One minute treadmill walking was done under the the following conditions with interactions in joint space:

\begin{itemize}
    \item No connection (NC). Both users walked in \emph{haptic transparency} mode with zero desired interaction torques.
    
    \item Bidirectional soft connection with zero neutral angle \\($\mt{K} = 30~\text{Nm/rad}, \mt{C} = 4~\text{Nms/rad}$).
    
    \item Bidirectional hard connection with zero neutral angle \\($\mt{K} = 70~\text{Nm/rad}, \mt{C} = 10~\text{Nms/rad}$).

    \item Bidirectional hard connection with $\mt{q}^\text{hip}_0 = 30^\circ$.

    \item Bidirectional hard connection with $\mt{q}^\text{knee}_0 = 20^\circ$.

    \item Unidirectional harder connection with zero neutral angle while only user A receives the visual feedback of the desired and actual position of his left ankle, and only user B feels the coupling force ($\mt{K}~=~100~\text{Nm/rad}, \mt{C}~=~10~\text{Nms/rad}$).
\end{itemize}

Treadmill speeds were set to 0.8 km/h. A relatively slow speed was chosen as a safety precaution during the connected trials. The treadmills of both users were manually started around the same time. For each of the conditions, the initial two gait cycles were not used in the analysis.

In the unidirectional condition, the desired ankle trajectory for user A was obtained by scaling up the displacement magnitude of the recorded ankle motion of a previous experienced user by 20\% along the vertical direction and 35\% along the horizontal direction in a hip-centered coordinate frame similar to the study of Marchal-Crespo et al. \cite{Marchal-Crespo2019-zo}. The desired trajectory was only for the left ankle, and no feedback was provided for the right leg. This resulted in an asymmetric desired gait pattern. The visual feedback is presented in Figure~\ref{fig:dyad_scheme}.

For the task space interaction, a preliminary test was conducted with both subjects stationary in single stance (no walking). Unidirectional spring and damper elements were rendered between the vertical positions ($\mt{K_y}~=~250~\text{N/m}, \mt{C_y}~=~50~\text{Ns/m}$) of the user's ankles. The leader (user A)'s robot rendered zero interaction torques while the follower (user B)'s robot rendered interaction torques determined by the unidirectional spring and damper elements. User A was asked to move his ankle vertically several times, and user B was asked to follow the forces he experienced. The users were not able to see each other during the task space interaction experiments.
\section{Results}
\label{sec:Results}

\subsection{Rendering Different Stiffness}

\begin{figure}[t!]
\centering
\vspace*{0.2cm}\includegraphics[width = 0.85\columnwidth]{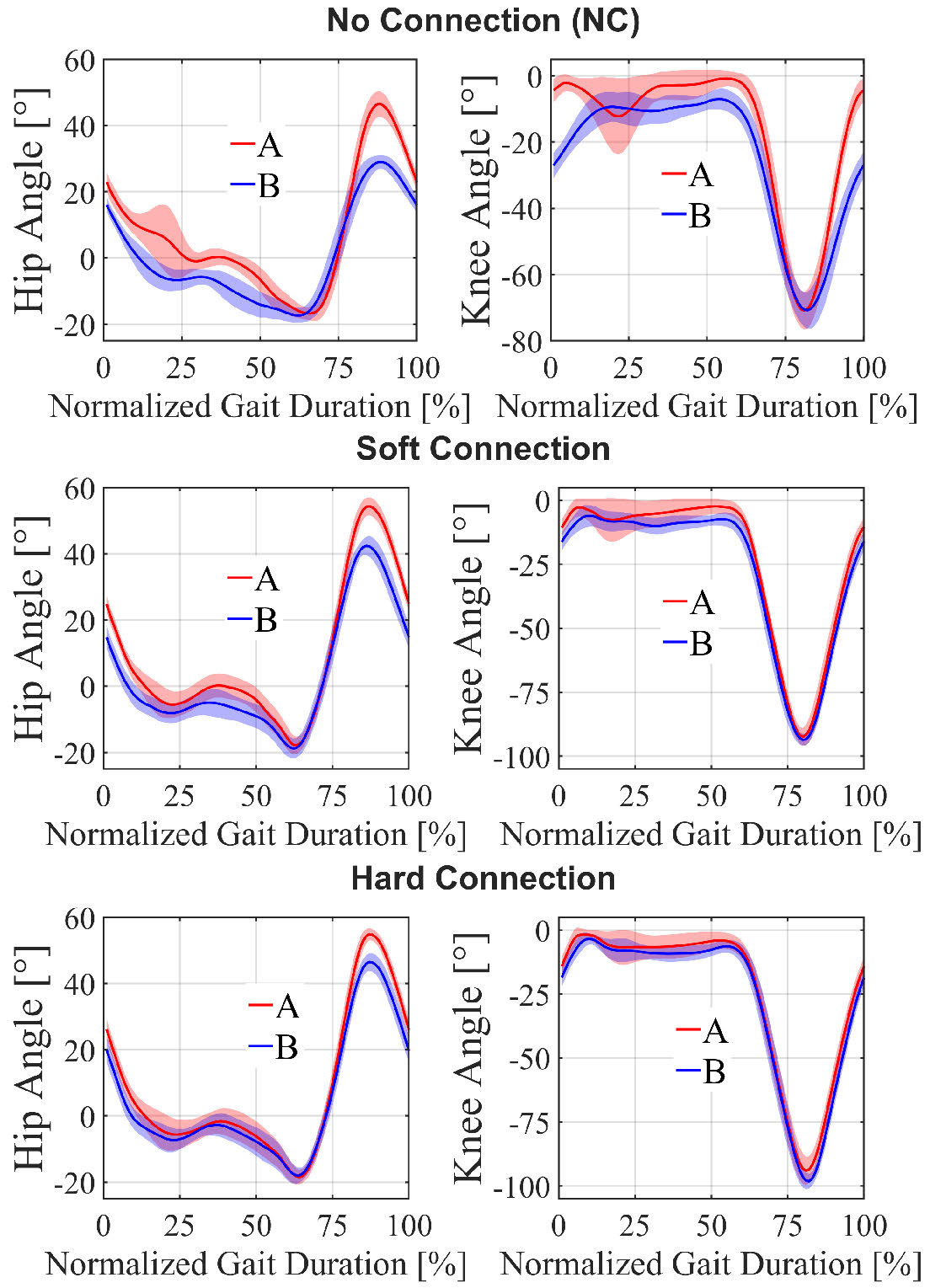}
\caption{Joint angles of users A and B with respect to normalized gait duration during different interaction properties. The data includes all steps of both legs during the one-minute trials. The shaded area represents $\pm$ one standard deviation. 
}
\label{fig:diff_stiff_vs_gait}
\end{figure}

Figure~\ref{fig:diff_stiff_vs_gait} shows the joint angles of the users during no connection (NC), soft interaction and hard interaction with zero neutral angles. Note that the NC plots are normalized with respect to the individual heel strikes of each user; the horizontal axes of the plot does not include any temporal correlation between the users in this non-connected trial.

During the NC condition, we observed that user B had a smaller range of motion for his hip and knee joints compared to user A. The peak angle difference between the subjects during the NC condition was around 20$^{\circ}$ for both the hip (at ~80\% of gait duration) and knee (at ~0\% of gait duration) joints. For the soft and hard interaction conditions, these differences decreased to ~12$^{\circ}$ and ~8$^{\circ}$, respectively, at the hip joints. For the knee joint, the connection resulted in ~6$^{\circ}$ and ~4$^{\circ}$ difference during the soft and hard interactions, respectively, at the heel strike instances. It is worth noting that the $x$-axes of the connected trials are normalized with respect to user A's heel strikes for both users. This shows that the connected trials also resulted in synchronization in the time domain.

Figure~\ref{fig:bar_plots} presents the mean absolute joint angle differences between user A and user B. It was seen that the average differences were 10$^{\circ}$, 6$^{\circ}$, and 3$^{\circ}$ at the hip joint for NC, soft and hard conditions, respectively. At the knee joint, average differences were observed as 11$^{\circ}$, 5$^{\circ}$ and 3$^{\circ}$.

\subsection{Changing the Neutral Angle}
\begin{figure}[t!]
\centering
\includegraphics[width = 0.85\columnwidth]{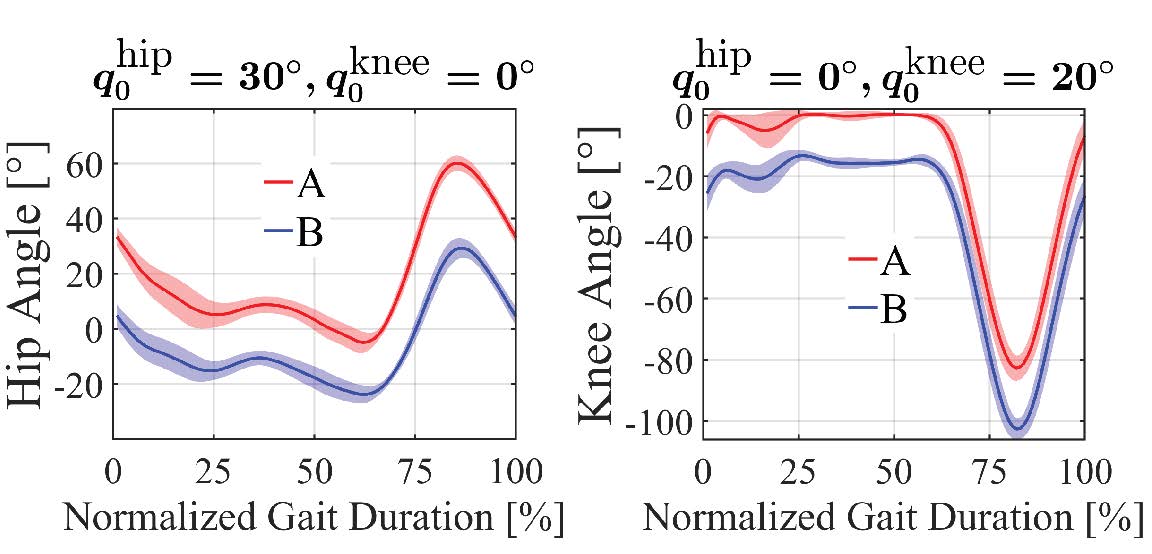}
\caption{Joint angles of users A and B with respect to normalized gait duration for virtual springs with different neutral angles. The data includes all steps of both legs during the one-minute trials. The shaded area represents $\pm$ one standard deviation.}
\label{fig:nonzero_neutral_vs_gait}
\end{figure}

\begin{figure}[t!]
\centering
\includegraphics[width = 0.9\columnwidth]{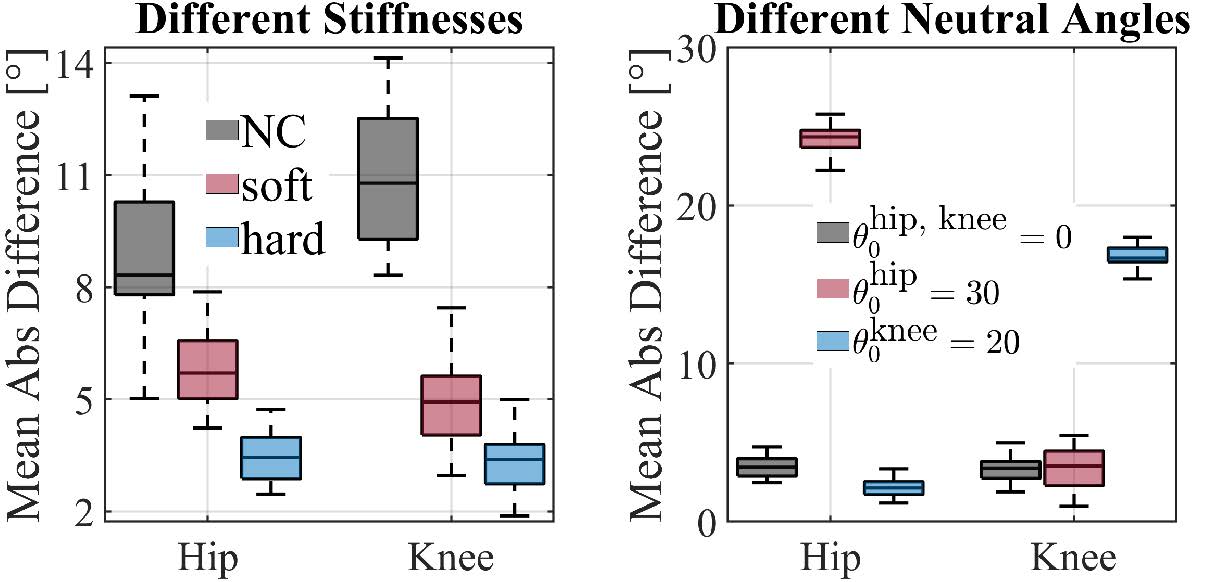}
\caption{Mean absolute joint angle difference between the users for trials with different stiffnesses (left) and neutral angles (right). Mean absolute values of the whole gait cycle for each leg are used as a single data point.}
\label{fig:bar_plots}
\end{figure}

Figure~\ref{fig:nonzero_neutral_vs_gait} shows the joint angles of each user with respect to the normalized gait duration for the non-zero neutral angle conditions. For the non-zero neutral angle at the hip, the hip joints of the users are offset from one another while their knee joints are aligned and synchronized. Similar behavior was also observed for the non-zero neutral angle at the knee as shown in Figure~\ref{fig:nonzero_neutral_vs_gait} and ~\ref{fig:bar_plots}. The right column of Figure~\ref{fig:bar_plots} shows the mean absolute difference in joint angles during these trials. The average difference in hip and knee joints was 24$^{\circ}$ and 17$^{\circ}$ for the trials with non-zero $q_0^{\text{hip}}$ and $q_0^{\text{knee}}$, respectively.

\vspace*{0.2cm}\subsection{Unidirectional Interaction}
\begin{figure}[t!]
\centering
\vspace*{0.2cm}\includegraphics[width = 0.9\columnwidth]{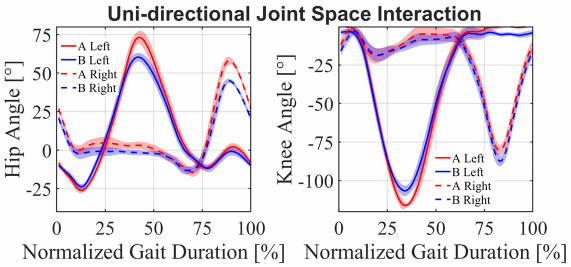}
\caption{Hip and knee joint angles during joint space unidirectional interaction trial. The data includes all steps of the one-minute trials. The shaded area represents $\pm$ one standard deviation.}
\label{fig:unilateral_vs_gait}
\end{figure}

Joint angles during the trials with joint space unidirectional interaction are presented in Figure~\ref{fig:unilateral_vs_gait}. Following the visually-displayed, scaled ankle trajectory resulted in an asymmetric gait for user A (leader, feels no interaction force). The difference between his left and right legs during peak flexion was around 15$^{\circ}$ and 36$^{\circ}$ for the hip and knee joints, respectively. The unidirectional interaction also resulted in an asymmetric gait for user B, even though he was not provided visual feedback. The difference between his left and right legs during peak flexion was around 15$^{\circ}$ and 30$^{\circ}$ for the hip and knee joints, respectively.

\begin{figure}[t!]
\centering
\includegraphics[width = 0.75\columnwidth]{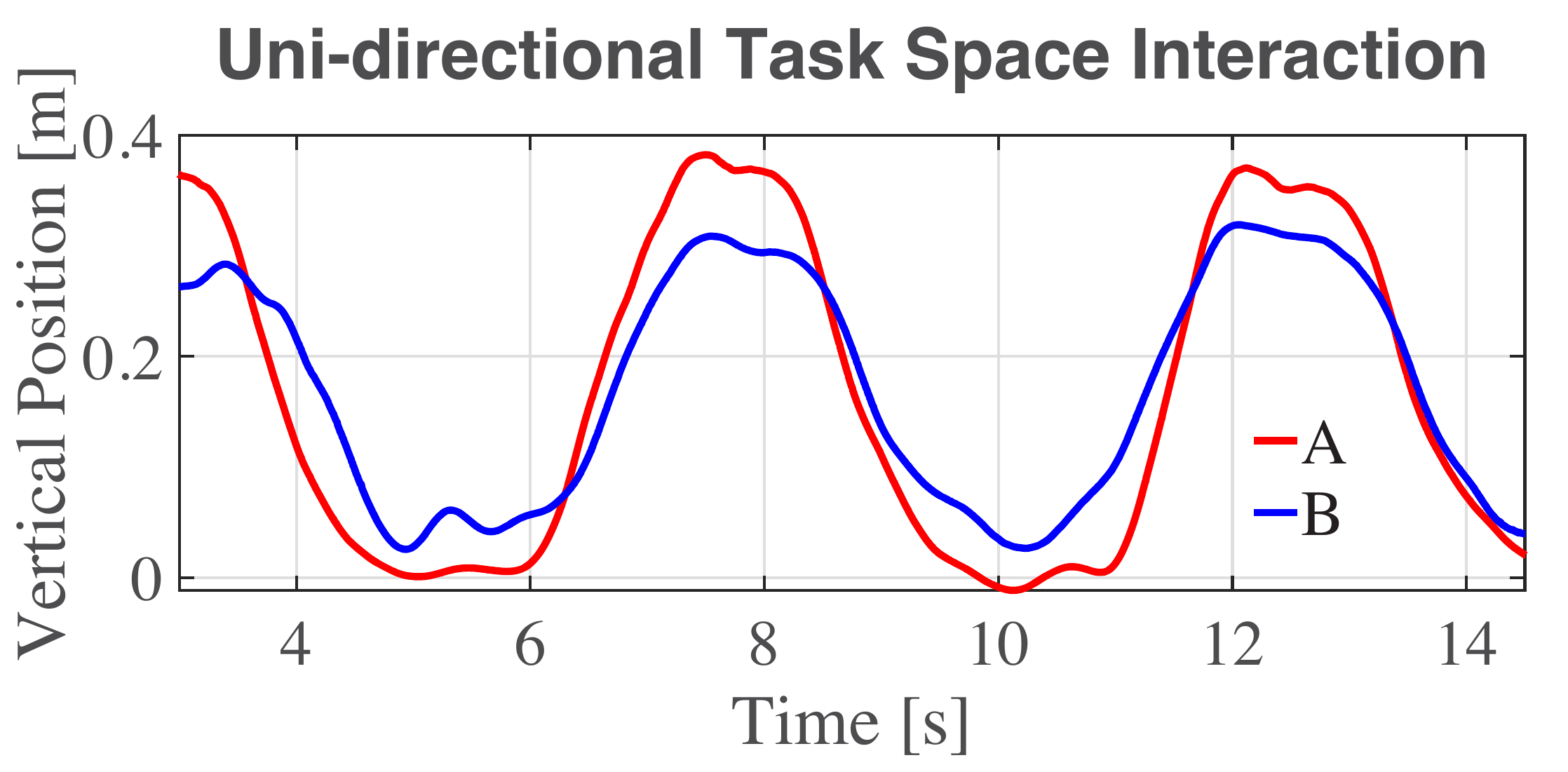}
\caption{Ankle vertical position with respect to time during the task space unidirectional interaction trial.}
\label{fig:task_sapce}
\end{figure}

Figure~\ref{fig:task_sapce} shows the vertical ankle positions of the users in the sagittal plane during the task space unidirectional interaction trial. The leader voluntarily moved his ankle vertically to a position of 40 cm, approximately. It was seen that the unidirectional force applied to user B moved his ankle to around 30 cm in the vertical direction.

\section{Discussion}
\label{sec:Discussion}

\subsection{Rendering Spring and Damper Elements with Different Characteristics}

During the walking trial without interaction, user B used a smaller range of motion for his knee and hip joints. This resulted in shorter, faster and unsynchronized steps with respect to user A. As the stiffness of the virtual connection increased during connected trials, the difference between each users' joint angles decreased, and their steps became more synchronized. In connected trials, knee joint angles of the users were closer to each other compared to their hip angles. This is because the same stiffness parameters were used for the hip and knee joints, and the hip joints must carry a larger weight than the knee joints. In future studies, larger stiffness parameters can be used for the hip joint to overcome this difference.

Rendering virtual springs with non-zero neutral angles resulted in the users' joints being pushed away from each other by an offset. This caused subjects to ``agree" on new gait patterns due to the forces they experienced. There was 6$^\circ$ and 3$^\circ$ difference on average between the neutral angles of the virtual springs and observed mean absolute difference for the hip and knee joints, respectively. This is due to the compliant nature of the rendered spring and damper elements. Increasing the stiffness would enforce the neutral angle more strictly while decreasing it would allow more deviation.

\begin{figure}[t!]
\centering
\vspace*{0.2cm}\includegraphics[width = 0.95\columnwidth]{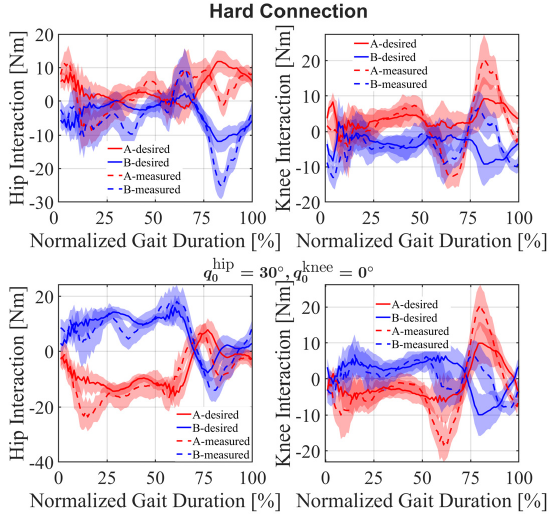}
\caption{Desired interaction torques commanded to WECC and actual interaction torques.}
\label{fig:intForce}
\end{figure}

Desired and realized interaction torques are presented in Figure~\ref{fig:intForce}. It can be seen that the difference between the desired and actual interaction torques increases near the end of the gait cycle (during swing). This is due to the fact that smaller control gains were used compared to~\cite{Kucuktabak2023haptic} as a safety precaution in these preliminary trials.

In the unidirectional trial, the leader (user A) demonstrated an asymmetric gait while following the visually-displayed, scaled ankle trajectory. User B was not provided the scaled desired trajectory. However, the proposed framework allowed the information of this modified gait pattern to be transmitted through the virtual physical coupling between the subjects.

\subsection{Implementation of Interaction Modes}

Physical interaction between two individuals can be divided mainly into three modes~\cite{Kucuktabak2021}: (1) collaboration: partners work together to achieve a common goal, (2) cooperation: partners have a common goal but different roles (e.g., teacher-student), (3) competition: partners can negatively affect one another in pursuit of their goals. 

In a clinical scenario, two impaired users can be provided with a visual display of a desired gait trajectory depending on their needs. Different interaction modes can be implemented by modifying the characteristics of the virtual coupling. For example, a bidirectional interaction can be used for a collaborative task where individuals help each other to better track the given trajectory. This interaction mode can be useful for the initial stages of rehabilitation, where users have limited ability due to their sensorimotor impairments. Unidirectional interaction can be used for a teacher-student scenario to demonstrate a desired trajectory through virtual physical interaction. Moreover, virtual springs with large neutral angles or negative stiffnesses can be used to generate conflicting goals between users for a competitive task. This interaction mode can be useful in the later stages of rehabilitation to promote motivation and higher exerted effort \cite{Basalp2021}. In future studies, we will use this framework to implement these interaction modes, investigating the effect of lower-limb haptic interaction on task performance and individual motor learning.

\subsection{Different Interaction Paradigms}

In the case of significant differences between the users' physical properties (i.e., weight, strength), symmetric bidirectional spring and damper elements might result in an unbalanced interaction in favor of the stronger user. In these situations, two sets of unidirectional spring and damper elements with different properties can be used to have a more balanced interaction (e.g., a stiffer spring attached to the stronger user and a softer spring attached to the weaker user).

In teleoperation, it is also common to create a two-port communication where the force on user A is transmitted to user B while the velocity of user B is transmitted to user A \cite{Nahri2021}. Such a virtual connection might be helpful in a scenario where a therapist can apply corrective forces to the patient while feeling how they move and their limitations. In future studies, these different interaction paradigms can also be implemented into this framework.

\subsection{Safety Considerations}

The forces due to the virtual coupling might cause unexpected disturbances, resulting in loss of balance, trembling or falls. Moreover, an unsafe movement on one side would be transferred to the other user through the virtual physical coupling. The WECC control running for both of the exoskeletons limits the maximum joint velocity, torque and power as constraints in the optimization scheme. However, it does not consider the balance of the user or the effects of the virtual coupling between two exoskeletons. Therefore, in future studies, it is necessary to implement higher-level safety precautions that would consider the balance of each user and the stability of the physical interaction.

\section{Conclusion}
\label{sec:Conclusion}
In this study, we developed an infrastructure to create virtual physical coupling between the lower-limbs of two individuals. The desired interaction torques are calculated and given to WECC control based on the properties of the virtual environment and instantaneous kinematics of the users.
\textcolor{black}{In this feasibility study, we tested the capability of this system to} render virtual environments with different properties (e.g., soft, hard, non-zero neutral angle), directionality (e.g., unidirectional, bidirectional), and interaction space (e.g., joint, task).
In the future, we will validate the capabilities of this infrastructure extensively with more users and implement additional safety features. Moreover, this framework will be used for the investigation how haptic interaction affects task performance and motor learning to improve motor outcomes of robot-mediated stroke rehabilitation.


\addtolength{\textheight}{-0cm}   

\bibliographystyle{bibliography/myIEEEtran} 
\bibliography{bibliography/references}

\begin{thebibliography}{10}
\providecommand{\url}[1]{#1}
\csname url@rmstyle\endcsname
\providecommand{\newblock}{\relax}
\providecommand{\bibinfo}[2]{#2}
\providecommand\BIBentrySTDinterwordspacing{\spaceskip=0pt\relax}
\providecommand\BIBentryALTinterwordstretchfactor{4}
\providecommand\BIBentryALTinterwordspacing{\spaceskip=\fontdimen2\font plus
\BIBentryALTinterwordstretchfactor\fontdimen3\font minus
  \fontdimen4\font\relax}
\providecommand\BIBforeignlanguage[2]{{%
\expandafter\ifx\csname l@#1\endcsname\relax
\typeout{** WARNING: IEEEtran.bst: No hyphenation pattern has been}%
\typeout{** loaded for the language `#1'. Using the pattern for}%
\typeout{** the default language instead.}%
\else
\language=\csname l@#1\endcsname
\fi
#2}}

\bibitem{Kucuktabak2021}
\BIBentryALTinterwordspacing
E.~B. K{\"u}{\c{c}}{\"u}ktabak, S.~J. Kim, Y.~Wen, K.~Lynch, and J.~L. Pons,
  ``Human-machine-human interaction in motor control and rehabilitation: a
  review,'' \emph{Journal of NeuroEngineering and Rehabilitation}, vol.~18,
  no.~1, p. 183, Dec 2021.
\BIBentrySTDinterwordspacing

\bibitem{Basalp2021}
E.~Basalp, P.~Wolf, and L.~Marchal-Crespo, ``Haptic training: Which types
  facilitate (re)learning of which motor task and for whom? answers by a
  review,'' \emph{IEEE Transactions on Haptics}, vol.~14, no.~4, pp. 722--739,
  2021.

\bibitem{Ganesh2014}
\BIBentryALTinterwordspacing
G.~Ganesh, A.~Takagi, R.~Osu, T.~Yoshioka, M.~Kawato, and E.~Burdet, ``Two is
  better than one: Physical interactions improve motor performance in humans,''
  \emph{Scientific Reports}, 2014.
\BIBentrySTDinterwordspacing

\bibitem{Takagi2017}
A.~Takagi, G.~Ganesh, T.~Yoshioka, M.~Kawato, and E.~Burdet, ``{Physically
  interacting individuals estimate the partner's goal to enhance their
  movements},'' \emph{Nature Human Behaviour}, vol.~1, no.~3, 2017.

\bibitem{Beckers2020}
\BIBentryALTinterwordspacing
N.~Beckers, E.~H.~F. van Asseldonk, and H.~van~der Kooij, ``Haptic human--human
  interaction does not improve individual visuomotor adaptation,''
  \emph{Scientific Reports}, vol.~10, no.~1, p. 19902, Nov 2020.
\BIBentrySTDinterwordspacing

\bibitem{Takagi2018}
\BIBentryALTinterwordspacing
A.~Takagi, F.~Usai, G.~Ganesh, V.~Sanguineti, and E.~Burdet, ``Haptic
  communication between humans is tuned by the hard or soft mechanics of
  interaction,'' \emph{{PLOS} Computational Biology}, vol.~14, no.~3, p.
  e1005971, Mar. 2018.
\BIBentrySTDinterwordspacing

\bibitem{Ivanova2022-ui}
E.~Ivanova, J.~Eden, G.~Carboni, J.~Kr{\"u}ger, and E.~Burdet,
  ``\BIBforeignlanguage{en}{Interaction with a reactive partner improves
  learning in contrast to passive guidance},''
  \emph{\BIBforeignlanguage{en}{Sci. Rep.}}, vol.~12, no.~1, p. 15821, Sept.
  2022.

\bibitem{Batson2020-wa}
J.~P. Batson, Y.~Kato, K.~Shuster, J.~L. Patton, K.~B. Reed, T.~Tsuji, and
  D.~Novak, ``\BIBforeignlanguage{en}{Haptic coupling in dyads improves motor
  learning in a simple force field},'' \emph{\BIBforeignlanguage{en}{Annu Int
  Conf IEEE Eng Med Biol Soc}}, vol. 2020, pp. 4795--4798, July 2020.

\bibitem{Diaz2011-bf}
I.~D{\'\i}az, J.~J. Gil, and E.~S{\'a}nchez,
  ``\BIBforeignlanguage{en}{Lower-limb robotic rehabilitation: Literature
  review and challenges},'' \emph{\BIBforeignlanguage{en}{J. robot.}}, vol.
  2011, pp. 1--11, 2011.

\bibitem{Kim2020}
S.~J. Kim, Y.~Wen, E.~B. Küçüktabak, S.~Zhan, K.~Lynch, L.~Hargrove, E.~J.
  Perreault, and J.~L. Pons, ``A framework for dyadic physical interaction
  studies during ankle motor tasks,'' \emph{IEEE Robotics and Automation
  Letters}, vol.~6, no.~4, pp. 6876--6883, 2021.

\bibitem{Kim2022}
S.~J. Kim, Y.~Wen, D.~Ludvig, E.~B. Küçüktabak, M.~R. Short, K.~Lynch,
  L.~Hargrove, E.~J. Perreault, and J.~L. Pons, ``Effect of dyadic haptic
  collaboration on ankle motor learning and task performance,'' \emph{IEEE
  Transactions on Neural Systems and Rehabilitation Engineering}, 2022.

\bibitem{B2018}
N.~Beckers, A.~Keemink, E.~van Asseldonk, and H.~van der Kooij, ``Haptic
  human-human interaction through a compliant connection does not improve motor
  learning in a force field,'' in \emph{Haptics: Science, Technology, and
  Applications}, D.~Prattichizzo, H.~Shinoda, H.~Z. Tan, E.~Ruffaldi, and
  A.~Frisoli, Eds.\hskip 1em plus 0.5em minus 0.4em\relax Cham: Springer
  International Publishing, 2018, pp. 333--344.

\bibitem{Middleton2015-ux}
A.~Middleton, S.~L. Fritz, and M.~Lusardi, ``\BIBforeignlanguage{en}{Walking
  speed: the functional vital sign},'' \emph{\BIBforeignlanguage{en}{J. Aging
  Phys. Act.}}, vol.~23, no.~2, pp. 314--322, Apr. 2015.

\bibitem{Bruni2018-bj}
M.~F. Bruni, C.~Melegari, M.~C. De~Cola, A.~Bramanti, P.~Bramanti, and R.~S.
  Calabr{\`o}, ``\BIBforeignlanguage{en}{What does best evidence tell us about
  robotic gait rehabilitation in stroke patients: A systematic review and
  meta-analysis},'' \emph{\BIBforeignlanguage{en}{J. Clin. Neurosci.}},
  vol.~48, pp. 11--17, Feb. 2018.

\bibitem{Kucuktabak2023haptic}
E.~B. K{\"u}{\c{c}}{\"u}ktabak, Y.~Wen, S.~J. Kim, M.~Short, D.~Ludvig,
  L.~Hargrove, E.~Perreault, K.~Lynch, and J.~Pons, ``Haptic transparency and
  interaction force control for a lower-limb exoskeleton,'' \emph{arXiv
  preprint arXiv:2301.06244}, 2023.

\bibitem{FongCanOpenDevelopment}
J.~Fong, E.~B. Küçüktabak, V.~Crocher, Y.~Tan, K.~Lynch, J.~Pons, and
  D.~Oetomo, ``{CANopen Robot Controller (CORC): An open software stack for
  human robot interaction development},'' \emph{The International Symposium on
  Wearable Robotics (WeRob2020)}, vol. 2020, 2020.

\bibitem{Marchal-Crespo2019-zo}
L.~Marchal-Crespo, P.~Tsangaridis, D.~Obwegeser, S.~Maggioni, and R.~Riener,
  ``\BIBforeignlanguage{en}{Haptic error modulation outperforms visual error
  amplification when learning a modified gait pattern},''
  \emph{\BIBforeignlanguage{en}{Front Neurosci}}, vol.~13, p.~61, Feb. 2019.

\bibitem{Nahri2021}
\BIBentryALTinterwordspacing
S.~N.~F. Nahri, S.~Du, and B.~J. Van~Wyk, ``A review on haptic bilateral
  teleoperation systems,'' \emph{Journal of Intelligent {\&} Robotic Systems},
  vol. 104, no.~1, p.~13, Dec 2021.
\BIBentrySTDinterwordspacing

\end{thebibliography}

\end{document}